# Fuzzy-based Navigation and Control of a Non-Holonomic Mobile Robot

Razif Rashid, I. Elamvazuthi, Mumtaj Begam, M. Arrofiq

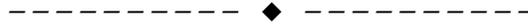

**Abstract**— In recent years, the use of non-analytical methods of computing such as fuzzy logic, evolutionary computation, and neural networks has demonstrated the utility and potential of these paradigms for intelligent control of mobile robot navigation. In this paper, a theoretical model of a fuzzy based controller for an autonomous mobile robot is developed. The paper begins with the mathematical model of the robot that involves the kinematic model. Then, the fuzzy logic controller is developed and discussed in detail. The proposed method is successfully tested in simulations, and it compares the effectiveness of three different set of membership of functions. It is shown that fuzzy logic controller with input membership of three provides better performance compared with five and seven membership functions.

—————————— ◆ ——————————

## 1 INTRODUCTION

IN recent years, there has been a growing interest in the design of feedback control laws for mobile robots with non-holonomic constraints [1-3]. Navigation of non-holonomic mobile robot is an interesting and challenging problem. Implementation is not simple because planning of actions to navigate robot requires high level of robot control skills. Robot as a system has two control variables (speed of left and right wheels), it's navigation is described with three coordinates (x, y and θ) and these facts make navigation problem as a challenge [4].

The development of fully autonomous control systems for mobile robots is made possible due to the increasing power of computational resources [5]. With these developments, many control schemes have been proposed to dealing with the control problem such as neural network [6-7] and fuzzy control [8-9]. Fuzzy logic-based techniques have been applied successfully to build the control system of autonomous intelligent mobile robots. Vamsi Mohan Peri et al. have developed a fuzzy logic controller to control the robot's motion along the predefined path [10]. M.K Singh et al. proposed a fuzzy control scheme consist of a heading angle between a robot and a special target [11].

Although recent research into mobile robots has progressed remarkably, there are still certain limits to accurate navigation. Fuzzy logic is especially advantageous for problems that cannot be easily represented by mathematical modeling because data is either unavailable, incomplete or the process is too complex. Other advantages of fuzzy logic-based techniques are that they allow building robust and smooth controllers starting from heuristic knowledge and qualitative models; considering imprecise, vague, and unreliable information; and integrating symbolic reasoning and numeric processing in the same framework [12].

In this paper, a new type of controller based on Fuzzy Logic for a non-holonomic mobile robot is proposed. The remainder of the paper is organized as follows: The kinematic model of a mobile robot is provided in section 2, and this is followed by the presentation of the materials and methods of the study in section 3. Result and Discussion of the controller is then given in section 4, and finally concluding remarks are given in section 5.

- *Razif Rashid is with the Department of Electrical and Electronic Engineering, Universiti Teknologi PETRONAS Bandar Seri Iskandar 31750 Tronoh.Perak, MALAYSIA.*
- *I. Elamvazuthi is with the Department of Electrical and Electronic Engineering, Universiti Teknologi PETRONAS Bandar Seri Iskandar 31750 Tronoh.Perak, MALAYSIA.*
- *Mumtaj Begam is with the Department of Electrical and Electronic Engineering, Universiti Teknologi PETRONAS Bandar Seri Iskandar 31750 Tronoh.Perak, MALAYSIA.*
- *M. Arrofiq is with the Department of Electrical and Electronic Engineering, Universiti Teknologi PETRONAS Bandar Seri Iskandar 31750 Tronoh.Perak, MALAYSIA.*

## 2 KINEMATICS MODEL OF MOBILE ROBOT

In order to analyze the system and develop controllers, a differential drive is considered for the kinematics models of mobile robot. The governing equations of the kinematic model taking non-holonomic constraints are well-known (see [10], [13], [14] and [15] for detail derivation). Figure 1 shows the variables used in the kinematic model and the equations discussed in the following:

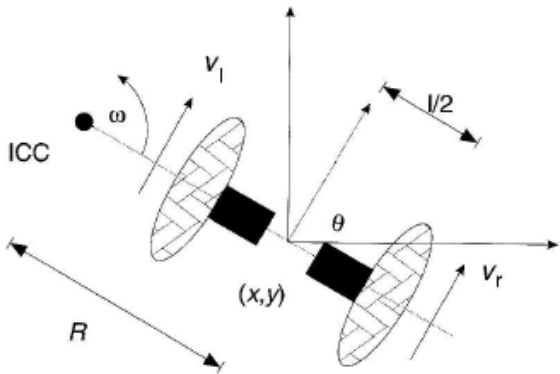

Figure 1. Kinematic model of the robot [6]

$V_r(t)$ = linear velocity of right wheel
$V_l(t)$ = linear velocity of left wheel
$\omega_r(t)$ = angular velocity of right wheel
$\omega_l(t)$ = angular velocity of left wheel
$r$ = nominal radius of each wheel
$L$ = distance between the two wheels
$R$ = instantaneous curvature radius of the robot trajectory, relative to the mid-point of the wheel axis
$ICC$ = Instantaneous Center of Curvature
$R - (L/2)$ = Curvature radius of trajectory described by left wheel
$R + (L/2)$ = Curvature radius of trajectory described by right wheel

With respect to ICC the angular velocity of the robot is given as follows.

$$\omega(t) = \frac{V_r(t)}{R + (\frac{L}{2})} \quad (1)$$

$$\omega(t) = \frac{V_l(t)}{R - (\frac{L}{2})} \quad (2)$$

$$\omega(t) = \frac{V_r(t) - V_l(t)}{L} \quad (3)$$

The instantaneous curvature radius of the robot trajectory relative to the mid-point of the wheel axis is given as

$$R = \frac{L(V_l(t) + V_r(t))}{2(V_l(t) - V_r(t))} \quad (4)$$

Therefore, the linear velocity of the robot is given as

$$V(t) = \omega(t)R = \frac{V_r(t) + V_l(t)}{2} \quad (5)$$

The kinematics equations in the world frame can be represented as follows.

$$X(t) = V(t)Cos\,\theta(t)$$
$$Y(t) = V(t)Sin\,\theta(t) \quad (6)$$
$$\theta(t) = \omega(t)$$

This implies

$$X(t) = \int_0^t V(t)\cos(\theta(t))dt$$
$$Y(t) = \int_0^t V(t)\sin(\theta(t))dt \quad (7)$$
$$\theta(t) = \int_0^t \omega(t)dt$$

The above equation can also be represented in the following form

$$\begin{bmatrix} V_x(t) \\ V_y(t) \\ \theta(t) \end{bmatrix} = \begin{bmatrix} \cos\theta & 0 \\ \sin\theta & 0 \\ 0 & 1 \end{bmatrix} \begin{bmatrix} V(t) \\ \omega(t) \end{bmatrix}$$

$$= \begin{bmatrix} V(t)\cos\theta \\ V(t)\sin\theta \\ \omega(t) \end{bmatrix}$$

$$= \begin{bmatrix} ((V_r + V_l)\cos\theta)/2 \\ ((V_r + V_l)\sin\theta)/2 \\ (V_r - V_l))/2 \end{bmatrix} \quad (8)$$

## 3 MATERIAL AND METHODS

### 3.1 Fuzzy Logic Controller

The objective is to control non-holonomic mobile robot using the Fuzzy Logic Controllers (FLC) to generate the velocities for both the right and the left motor of the mobile robot, which allows the robot to move from path to other path, similar to [10]. The FLC used has two input: error in the position and error in the angle of the robot. Thus, the FLC is two inputs, two output system. Figure 2 shows the robot in Cartesian space and Figure 3 shows the robot in Cartesian space during its motion.

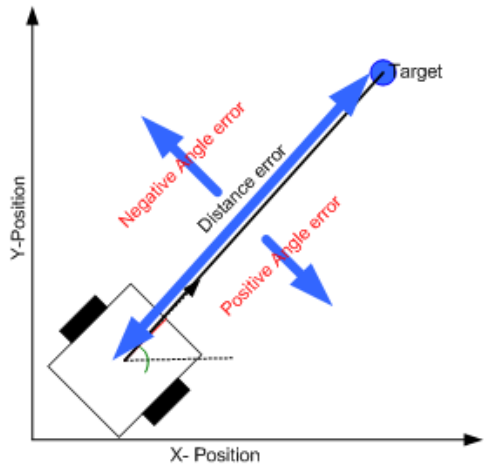

Figure 2. Robot in Cartesian space

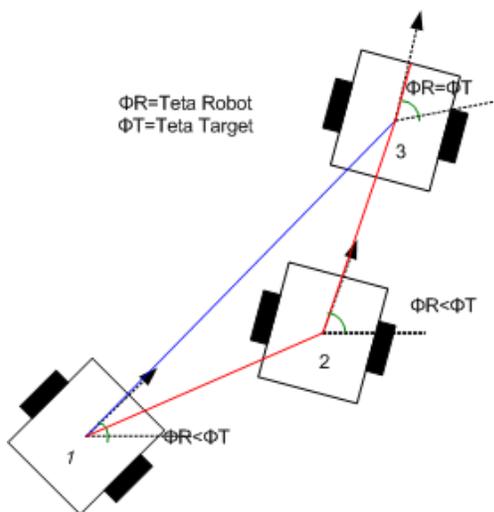

Figure 3. Robot in Cartesian space during running

The output of the fuzzy control would be pulse-width-modulated signal to control are angular velocities of the left and right motors of the mobile robot. Figure 4 shows the block diagram of mobile robot system .

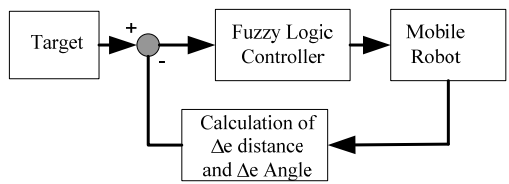

Figure 4. Block diagram of mobile robot system

### 3.2 Algorithm

The fuzzy logic algorithm is as follows [16]:
Step 1: Define the linguistic variable(s) for input and output system.
- Compute two input variables: angle error (the different angle between goal angle and robot's heading) and distance error (the difference between the current position and goal position).
- Compute two output variables: the velocities of the left and right motors.

Step 2: Define fuzzy set
- The fuzzy set value of the fuzzy variables is as shown in Table 1.
- The fuzzy set value is set of overlapping values represented by triangular shape that is called the fuzzy membership function

**Table 1. Notations for the Fuzzy Logic Input**

| Angle : $\theta$ error (e theta) | Distance : error (ed) |
|---|---|
| VSN:Very Small Negative | Z: Zero |
| SN : Small Negative | N : Near |
| N : Negative | VN: Very Near |
| Z: Zero | M: Middle |
| P : Positive | F: far |
| BP : Big Positive | VF: Very Far |
| VBP: Very Big Positive | VBF:Very Big Far |

Step 3: Define Fuzzy rules
- The operation of the system utilizes the fuzzy rule.

Step 4: Defuzzification
- The output action given by the input conditions, the centroid defuzzification computes the velocity output of the fuzzy controller.

### 3.3 Simulation

Simulation was conducted using the simulink model of the mobile robot as shown in Figure 5.

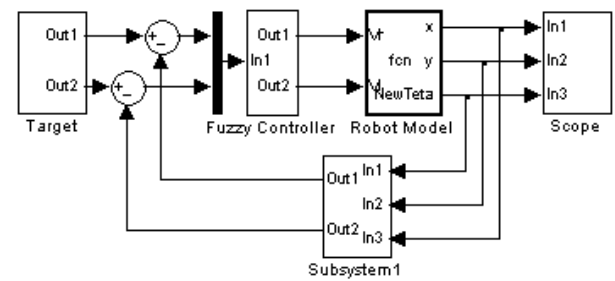

Figure 5. Simulink model of the mobile robot

## 4 RESULT & DISCUSSION

### 4.1 Fuzzy Representations

The output notation for making fuzzy rules is: Fast (F), Medium (M) and Slow (S). Fuzzy logic rules using three membership functions for right motor is shown in the Table 2 and left motor in shown in Table 3.

**Table 2. Fuzzy rule for velocity of the right motor**

| Δℓt /Δℓd | F | M | Z |
|---|---|---|---|
| N | M | M | S |
| Z | F | M | S |
| P | F | F | F |

**Table 3. Fuzzy rule for velocity of the left motor**

| Δℓt /Δℓd | F | M | Z |
|---|---|---|---|
| N | F | F | M |
| Z | F | M | S |
| P | M | M | S |

Example of Fuzzy rule for three membership of the differential drive mobile robot are shown below:

- R1: IF ℓt is Negative angle (N) and ℓd is distance Far (F) then the velocity of the right motor is Medium (M) and the velocity of the left motor is Fast (F).
- R2: IF ℓt is Negative angle (N) and ℓd is distance Medium (M) then the velocity of the right motor is Medium (M) and the velocity of the left motor is Fast (F).

The representation of the input fuzzy membership function of the angle error, the distance error , the representation of the output fuzzy membership function velocity of the right and left motors for three membership functions are shown in Figure 6 to 9.

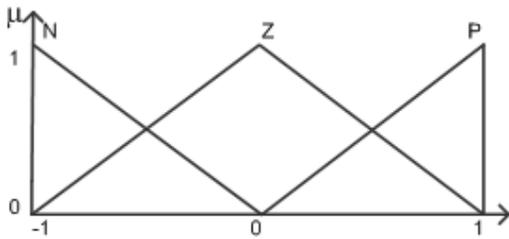

Figure 6 Representation of the input fuzzy three membership function of the angle error

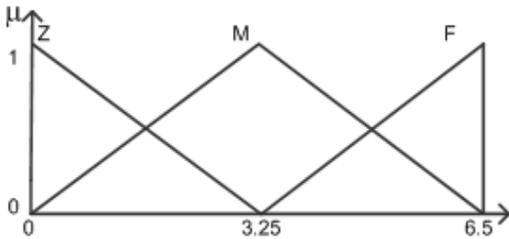

Figure 7 Representation of the input fuzzy three membership function of the distance error

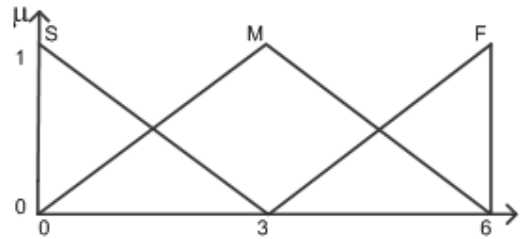

Figure 8 Representation of the output of the fuzzy three membership function velocity of the Right motor

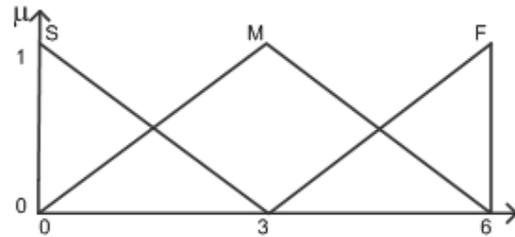

Figure 9 Representation of the output of the fuzzy three membership function velocity of the left motor

The output notation for making fuzzy rules is: Very Fast (VF), Fast (F), Medium (M), Slow (S) and Very Slow (VS). The notation for marking fuzzy rules using Five membership function for right motor is shown in the Table 4 and left motor in show in Table 5.

**Table 4. Fuzzy rule for velocity of the right motor**

| Δℓd /Δℓt | VF | F | M | N | Z |
|---|---|---|---|---|---|
| SN | M | S | VS | VS | VS |
| N | F | M | S | VS | VS |
| Z | VF | F | M | S | VS |
| P | VF | F | M | S | S |
| BP | VF | F | F | M | M |

**Table 5. Fuzzy rule for velocity of the left motor**

| Δℓd /Δℓt | VF | F | M | N | Z |
|---|---|---|---|---|---|
| SN | VF | F | F | M | M |
| N | VF | F | M | S | S |
| Z | VF | F | M | S | VS |
| P | F | M | S | VS | VS |
| BP | M | S | VS | VS | VS |

Example of Fuzzy rule for five membership of the differential drive mobile robot are shown below:

- R1: IF ℓt is Small Negative angle (SN) and ℓd is distance Very Far (VF) then the velocity of the right motor is Medium (M) and the velocity of the left motor is Very Fast (VF).
- R2: IF ℓt is Small Negative angle (SN) and ℓd is distance Far (F) then the velocity of the right motor is Slow (S) and the velocity of the left motor is Fast (F).

The representation of the input fuzzy membership function of the angle error, the distance error, the representation of the output fuzzy membership function velocity of the right and left motors for five membership functions are shown in Figure 10 to 13.

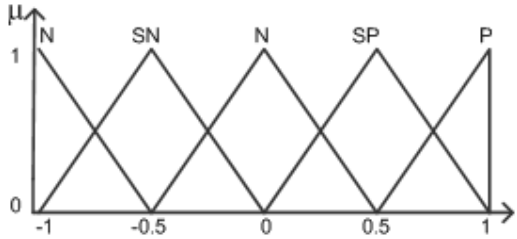
Figure 10 Representation of the input fuzzy five membership function of the angle error

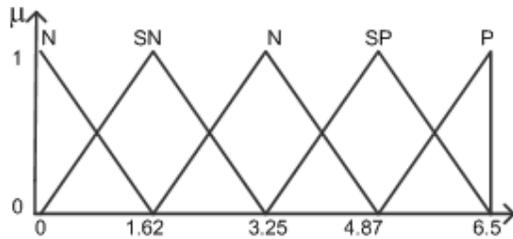
Figure 11 Representation of the input fuzzy five membership function of the distance error

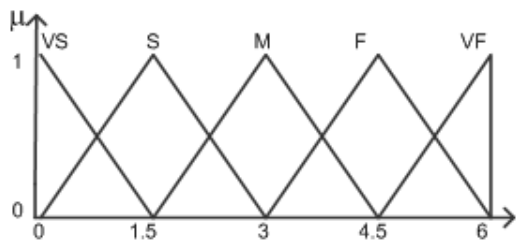
Figure 12 Representation of the output of the fuzzy five membership function velocity of the Right motor

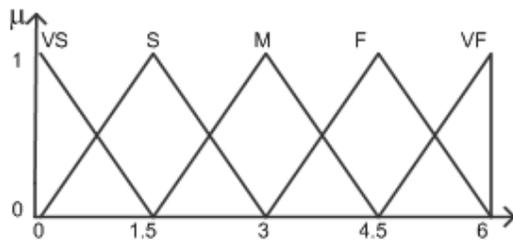
Figure 13 Representation of the output of the fuzzy five membership function velocity of the left motor

The output notation for making fuzzy rules is: Very Fast1 (VF2), Very Fast (VF1), Fast (F), Medium (M), Slow (S), Very Slow 1 (VS1) and Very Slow 2 (VS2). The notation for marking fuzzy rules using Seven membership function for right motor is shown in the Table 6 and left motor in show in Table 7.

Table 6. Fuzzy rule for velocity of the right motor

| Δℓd /Δℓt | VBP | VF | F | M | N | VNZ | Z |
|---|---|---|---|---|---|---|---|
| VSN | M | F | S | VS1 | VS1 | VS2 | VS2 |
| SN | F | M | S | VS1 | VS1 | VS1 | VS2 |
| N | VF1 | F | M | S | VS1 | VS1 | VS2 |
| Z | VF2 | VF1 | F | M | S | VS1 | VS2 |
| P | VF2 | VF1 | F | M | S | S | VS1 |
| BP | VF2 | VF1 | F | F | M | M | S |
| VBP | VF2 | VF2 | VF | F | M | M | S |

Table 7. Fuzzy rule for velocity of the left motor

| Δℓd /Δℓt | VBP | VF | F | M | N | VNZ | Z |
|---|---|---|---|---|---|---|---|
| VSN | VF2 | VF2 | VF | F | M | M | S |
| SN | VF2 | VF1 | F | F | M | M | S |
| N | VF2 | VF1 | F | M | S | S | VS1 |
| Z | VF2 | VF1 | F | M | S | VS1 | VS2 |
| P | VF1 | F | M | S | VS1 | VS1 | VS2 |
| BP | F | M | S | VS1 | VS1 | VS1 | VS2 |
| VBP | M | F | S | VS1 | VS1 | VS2 | VS2 |

Example of Fuzzy rule for seven membership of the differential drive mobile robot are show below:

- R1: IF ℓt is Very Small Negative angle (VSN) and ℓd is distance Very Big Positive (VBP) then the velocity of the right motor is Medium (M) and the velocity of the left motor is Very Fast 2 (VF2).
- R2: IF ℓt is Very Small Negative angle (VSN) and ℓd is distance Very Far (VF) then the velocity of the right motor is Fast (F) and the velocity of the left motor is Very Fast 2 (VF2).

The representation of the input fuzzy membership function of the angle error, the distance error, the representation of the output fuzzy membership function velocity of the right and left motors for five membership functions are shown in Figure 14 to 17.

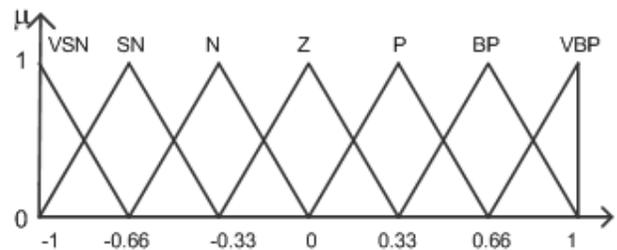
Figure 14 Representation of the input fuzzy seven membership function of the angle error

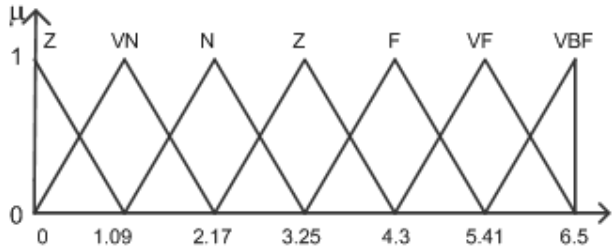

Figure 15 Representation of the input fuzzy seven membership function of the distance error

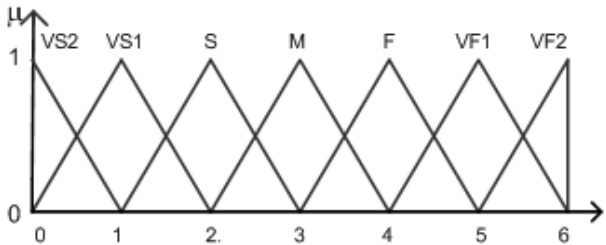

Figure 16 Representation of the output of the fuzzy seven membership function velocity of the right motor

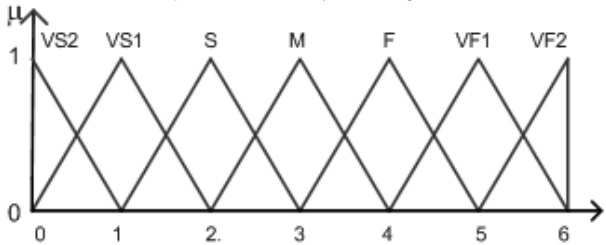

Figure 17 Representation of the output of the fuzzy seven membership function velocity of the left motor

### 4.2 System Response

The plot in figures 18, 19, 20, and 21 compare the simulation result obtained of the response of the robot for three, five and seven membership functions of fuzzy logic controller. Fig. 18 shows the angle of orientation of the robot as a function of time. Fig. 19 shows the x position of the mobile robot as a function of time. Fig. 20 shows the y position of the mobile robot as a function of time. Fig. 21 shows the position of mobile robot respect frame x position and y position.

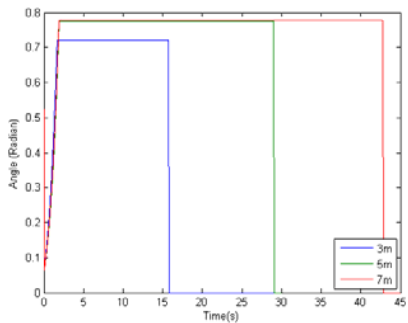

Figure 18. Angle of orientation as a function of time

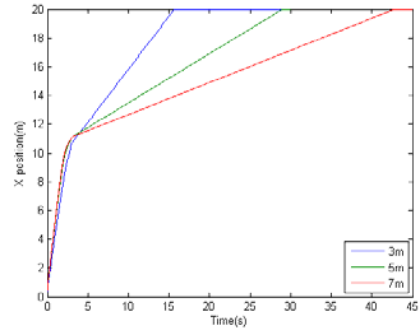

Figure 19. X-Position as a function of time

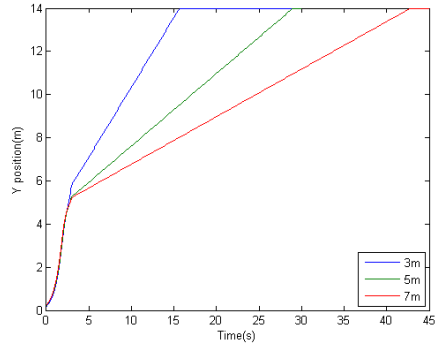

Figure 20. Y-Position as a function of time

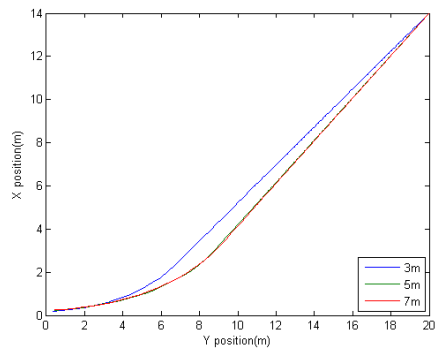

Figure 21. Position of mobile robot

The results shows that the three membership function has the best performance, followed by five and seven for angle of orientation, x position, y position and xy position as a whole. The plot in figures 22, 23, and 24 shows the velocity profile as a function of time for the robot for three, five and seven membership functions respectively.

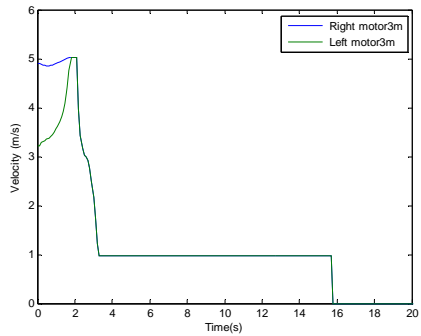

Figure 22. Velocity of mobile robot for three membership function

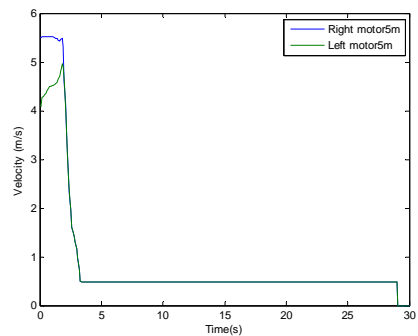

Figure 23. Velocity of mobile robot for five membership function

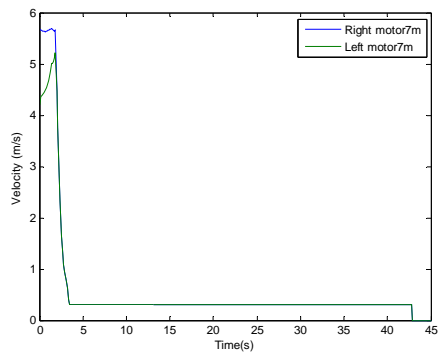

Figure 24. Velocity of mobile robot for seven membership function

The results suggest that the three membership function has the fastest response, followed by five and seven for the velocity of the robot. Table 8 provides the performance comparison of three, five and seven membership functions for the time to arrive at target and time for robot angle with constant parameters for the distance, sampling time.

**Table 8. Performance evaluation of difference membership function of autonomous mobile robot**

| Membership Function / Parameters | Three | Five | Seven |
|---|---|---|---|
| Distance | 24.41m | 24.41m | 24.41m |
| Sampling time | 0.1s | 0.1s | 0.1s |
| Time achieve target | 15.8s | 29.1s | 42.9s |
| Time angle robot=Time angle Target | 0.18s | 0.20s | 0.20s |
| Total rules | 9 | 25 | 49 |

Comparing the results from Table 8, it can be seen that the time to arrive at target is fastest by three, followed by five and seven. A similar trend is also observed for time for robot angle. It can be deduced that the best performance of the fuzzy logic controller is based on the least number of membership functions.

## 5 CONCLUSION

This paper has presented the complex and nonlinear problem such as the mobile robot navigation that can be solved by relative non-complex fuzzy technique. The proposed method is successfully tested in simulations, and it compares the effectiveness of three different set of membership of functions. It is shown that fuzzy logic controller with input membership of three provides better performance compared with five and seven membership functions. In future, a hybrid technique involving fuzzy and neural networks system could be explored to enhance the performance.

### ACKNOWLEDGMENT

The authors would like to thank Universiti Teknologi PETRONAS for supporting this work.

### REFERENCES

[1] Dixon, W.E., Jiang, Z.P., and Dawson, D.M., Global exponential setpoint control of wheeled mobile robots: a Lyapunov approach, Automatica, 2000, 36, pp. 1741–1746
[2] Duleba, I., and Sasiaek, J., Calibration of controls in steering nonholonomic systems, Control Eng. Pract., 2001.
[3] Dongbing, G., and Huosheng, H., Receding horizon tracking control of wheeled mobile robots, IEEE Trans. Control Syst. Technol., 2006, 14, (4), pp. 743–749
[4] Yue Liyong, Xie Wei, An adaptive tracking method for monholonomic wheeled mobile robots, Proceedings of the 26th Chinese Control Conference July 26-31, 2007, Zhangjiajie, Hunan, China

 

[5] W. Ren, J.-S. Sun, R. Beard and T. McLain, Experimental validation of an autonomous control system on a mobile robot platform, IET Control Theory Appl., 2007, 1, (6), pp. 1621–1629.

[6] Oubbati M, Schanz M, Levi P., Kinematic and Dyanamic Adaptive Control of a Nonholonomic Mobile Robot using a RNN[C], Proceedings 2005 IEEE International Symposium on Computatianal intelligence in Robotics and Automation, 2005: 27-33.

[7] Das T, Kar I N, Chaudhury S., Simple neuron-based adaptive controller for a nonholonomic mobile robot including actuator dynamics[J], Neuracomputing, 2006, 69: 2140-2151.

[8] Maalouf E, Saad M, Soliah H., A higher level path tracking controller for a four-wheer differentially steered mobile robot[J], Robotics and Autonomous Systems, 2006, 54: 23-33.

[9] Das T, Kar I N., Design and implementation of an adaptive fuzzy logic-based controller for wheeled mobile robots [J]. IEEE Trans. Control Systems Technology, 2006, 14(3): 501-510.

[10] V.M Peri, D Simon, Fuzzy Logic controller for an autonomous robot", Fuzzy Information Processing Society, 2005. NAFIPS 2005. Annual Meeting of the North American, ISBN: 0-7803-9187-X, 337- 342.

[11] M.K Singh, D.R.Parhi " Intelligent controller for mobile robot: Fuzzy logic approach" IACMAG 2008.

[12] Iluminada, B., Francisco J. M.V, Víctor, B and Joaquín, F., Design of embeded DSP-based fuzzy controllers for autonomous mobile robots, IEEE Transactions on Industrial Electronics, Vol. 55, No. 2, February 2008.

[13] Dudek and Jenkin, Computational principles of mobile robotic, Cambridge University Publications, 2000.

[14] M.I Rieiro and P.Lima, Kinematic models of mobile robots".http://omni.isr.ist.utl.pt/~mir/cadeiras/robmovel/kinematics.pdf

[15] http://www.robotix.gr/robotics/XBotPart2/

[16] R.Choomuang &N. Afzulpurkar/ Hybrid Kalman Filter/Fuzzy Logic based position control of autonomous mobile robot, pp. 197 - 208, International Journal of Advanced Robotic Systems, Volume 2, Number 3 (2005), ISSN 1729-8806.

**Razif Rashid i**s with the Department of Electrical and Electronic Engineering of Universiti Teknologi PETRONAS (UTP), Malaysia. His research interests include Robotics and Mechatronics.

**I. Elamvazuthi** is with the Department of Electrical and Electronic Engineering of Universiti Teknologi PETRONAS (UTP), Malaysia.

**Mumtaj Begam** is with the Department of Electrical and Electronic Engineering of Universiti Teknologi PETRONAS (UTP), Malaysia.

**M. Arrofiq** is with the Department of Electrical and Electronic Engineering of Universiti Teknologi PETRONAS (UTP), Malaysia.